# A Family of Computationally Efficient and Simple Estimators for Unnormalized Statistical Models

**Miika Pihlaja**
Dept of Mathematics & Statistics,
Dept of Comp Sci and HIIT
University of Helsinki
miika.pihlaja@helsinki.fi

**Michael Gutmann**
Dept of Comp Sci, HIIT and
Dept of Mathematics & Statistics,
University of Helsinki
michael.gutmann@helsinki.fi

**Aapo Hyvärinen**
Dept of Mathematics & Statistics,
Dept of Comp Sci and HIIT
University of Helsinki
aapo.hyvarinen@helsinki.fi

## Abstract

We introduce a new family of estimators for unnormalized statistical models. Our family of estimators is parameterized by two nonlinear functions and uses a single sample from an auxiliary distribution, generalizing Maximum Likelihood Monte Carlo estimation of Geyer and Thompson (1992). The family is such that we can estimate the partition function like any other parameter in the model. The estimation is done by optimizing an algebraically simple, well defined objective function, which allows for the use of dedicated optimization methods. We establish consistency of the estimator family and give an expression for the asymptotic covariance matrix, which enables us to further analyze the influence of the nonlinearities and the auxiliary density on estimation performance. Some estimators in our family are particularly stable for a wide range of auxiliary densities. Interestingly, a specific choice of the nonlinearity establishes a connection between density estimation and classification by nonlinear logistic regression. Finally, the optimal amount of auxiliary samples relative to the given amount of the data is considered from the perspective of computational efficiency.

## 1 INTRODUCTION

It is often the case that the statistical model related to an estimation problem is given in unnormalized form. Estimation of such models is difficult. Here we derive a computationally efficient and practically convenient family of estimators for such models.

The estimation problem we try to solve is formulated as follows. Assume we have a sample of size $N_d$ of a random vector $\mathbf{x} \in \mathbb{R}^n$ with distribution $p_d(\mathbf{x})$. We want to estimate a parameterized model

$$p_m(\mathbf{x}; \varphi) = \frac{p_m^0(\mathbf{x}; \varphi)}{Z(\varphi)}, \quad Z(\varphi) = \int p_m^0(\mathbf{x}; \varphi) \, d\mathbf{x} \quad (1)$$

for the data density. Here $p_m^0(\mathbf{x}; \varphi)$ is the unnormalized model, which specifies the functional form of the density, and $Z(\varphi)$ is the normalizing constant (partition function). Our paper deals with estimating the parameters $\varphi$ when the evaluation of the normalizing constant is unfeasible. Many popular models such as Markov random fields (Roth and Black, 2009; Köster et al., 2009) and multi-layer networks (Osindero et al., 2006; Köster and Hyvärinen, 2010) face this problem.

Classically, in Maximum Likelihood Estimation (MLE), it is necessary to have an analytical expression for the normalizing constant $Z(\varphi)$. For that reason it cannot be used to estimate unnormalized models. If an analytical expression is not available, Monte Carlo methods can be used to evaluate $Z(\varphi)$ (Geyer and Thompson, 1992; Hinton, 2002). Another option is to maximize alternative objective functions (Besag, 1974; Hyvärinen, 2005; Gutmann and Hyvärinen, 2010).

Here we propose in the same vein a whole family of objective functions to estimate unnormalized models. A particular instance of the family is closely related to Maximum Likelihood Monte Carlo (Geyer and Thompson, 1992), and we will see that the family includes Noise Contrastive Estimation (Gutmann and Hyvärinen, 2010) as a special case. The paper is structured as follows. We start by defining our estimator family and stating some basic properties in section 2. We then discuss how to choose particular instances from the family of estimators in section 3. We validate the theoretical results with simulations in section 4. Section 5 concludes the paper.

## 2  THE NEW ESTIMATOR FAMILY

First we motivate the definition of the new estimator family by formulating Maximum Likelihood Estimation as a variational problem. After formally defining the family, we establish some properties, such as consistency and asymptotic normality.

### 2.1  MAXIMUM LIKELIHOOD AS VARIATIONAL PROBLEM

Maximizing likelihood is equivalent to minimizing the Kullback-Leibler divergence between the data and the model densities, under the constraint that the latter is properly normalized.[1] We can use Lagrange multipliers to impose the normalization constraint, giving us the objective functional

$$\tilde{J}_{ML}[p_m^0] = \int p_d(\mathbf{x}) \log p_m^0(\mathbf{x})\, d\mathbf{x} - \lambda \left( \int p_m^0(\mathbf{x})\, d\mathbf{x} - 1 \right),$$

where $\lambda$ is a Lagrange multiplier. Determining the optimal value of $\lambda$ requires integration over the model density, which corresponds to evaluating the partition function $Z(\varphi)$. We can avoid that by introducing a new objective functional with auxiliary density $p_n$, which takes as an argument the log-model density $f$

$$\tilde{J}[f] = \int p_d \log \exp(f) - \int p_n \frac{\exp(f)}{p_n}. \quad (2)$$

Taking now the variational derivative with respect to $f$, we get

$$\delta \tilde{J}[f] = p_d - \exp(f) \quad (3)$$

which shows that the only stationary point is given by $f = \log p_d$. Note that in contrast to the case of MLE above, where the search is restricted to the space of functions integrating to one, here we optimize over the space of arbitrary sufficiently smooth functions $f$.

### 2.2  DEFINITION OF THE ESTIMATOR FAMILY

We propose to replace logarithm and identity by two nonlinear functions $g_1(\cdot)$ and $g_2(\cdot)$ defined in $\mathbb{R}_+$ and taking values in $\mathbb{R}$. This gives us the following family of objective functionals

$$\tilde{J}_g[f] = \int p_d\, g_1\left(\frac{\exp(f)}{p_n}\right) - \int p_n\, g_2\left(\frac{\exp(f)}{p_n}\right). \quad (4)$$

Calculation of the functional derivatives shows that the nonlinearities must be related by

$$\frac{g_2'(q)}{g_1'(q)} = q \quad (5)$$

---
[1]In what follows we often omit the explicit arguments of the densities for clarity, writing $p_m, p_n$ and $p_d$. In this case the integrals are taken over $\mathbb{R}^n$.

in order obtain $f = \log p_d$ as the unique stationary point.

In practical estimation tasks we use a parameterized model and compute the empirical expectations over the data and auxiliary densities using samples $(\mathbf{x}_1, \mathbf{x}_2, \ldots, \mathbf{x}_{N_d})$ and $(\mathbf{y}_1, \mathbf{y}_2, \ldots, \mathbf{y}_{N_n})$ from $p_d$ and $p_n$ respectively, where $\mathbf{x}_i, \mathbf{y}_j \in \mathbb{R}^n$. We also include the negative log-partition function as an additional parameter $c$, giving us the model

$$\log p_m(\mathbf{u};\theta) = \log p_m^0(\mathbf{u};\varphi) + c, \quad \theta = \{\varphi, c\}. \quad (6)$$

Note that $p_m(\mathbf{u};\theta)$ will only integrate to one for some particular values of the parameters. This leads to the following sample version of the objective function in (4)

$$J_g(\theta) = \frac{1}{N_d}\sum_{i=1}^{N_d} g_1\left(\frac{p_m(\mathbf{x}_i;\theta)}{p_n(\mathbf{x}_i)}\right) - \frac{1}{N_n}\sum_{j=1}^{N_n} g_2\left(\frac{p_m(\mathbf{y}_j;\theta)}{p_n(\mathbf{y}_j)}\right). \quad (7)$$

We define our estimator $\hat{\theta}_g$ to be the parameter value that maximizes this objective function.

### 2.3  PROPERTIES OF THE ESTIMATOR FAMILY

In this section, we will show that our new estimator family is consistent and asymptotically normally distributed. We will also provide an expression for the asymptotic covariance matrix which, as we will see, depends on the choice of $g_1(\cdot), g_2(\cdot)$ and $p_n$. This gives us a criterion to compare different estimators in the family.[2]

**Theorem 1.** (Non-parametric estimation) *Let $g_1(\cdot)$ and $g_2(\cdot)$ be chosen to satisfy $g_2'(q)/g_1'(q) = q$. Then $\tilde{J}_g(f)$ has a stationary point at $f(\mathbf{u}) = \log p_d(\mathbf{u})$. If furthermore $g_1(\cdot)$ is strictly increasing, then $f(\mathbf{u}) = \log p_d(\mathbf{u})$ is a maximum and there are no other extrema, as long as the auxiliary density $p_n(\mathbf{u})$ is chosen so that it is nonzero wherever $p_d(\mathbf{u})$ is nonzero.*

We can further show that this result carries over to the case of parametric estimation with sample averages from $p_d$ and $p_n$. Using a parameterized model, we restrict the space of functions where the true density of the data is searched for. Thus, we will make the standard assumption that the data density is included in the model family, i.e. there exists $\theta^\star$ such that $p_d = p_m(\theta^\star)$.

**Theorem 2.** (Consistency) *If conditions 1.-4. hold, then $\hat{\theta}_g \xrightarrow{P} \theta^\star$.*

1. $p_n$ is nonzero whenever $p_d$ is nonzero

---
[2]Proofs of the following theorems are omitted due to the lack of space.

2. $g_1(\cdot)$ and $g_2(\cdot)$ are strictly increasing and satisfy $g_2'(q)/g_1'(q) = q$

3. $\sup_\theta |J_g(\theta) - J_g^\infty(\theta)| \xrightarrow{P} 0$

4. Matrix $\mathcal{I} = \int p_d(\mathbf{u}) \psi(\mathbf{u}) \psi(\mathbf{u})^T g_2'\left(\frac{p_d(\mathbf{u})}{p_n(\mathbf{u})}\right) d\mathbf{u}$ is full rank, and $p_n$ and $g_2(\cdot)$ are chosen such that each of the integrals corresponding to the elements of the matrix is finite.

Here we define $\psi(\mathbf{u}) = \nabla_\theta \log p_m(\mathbf{u}, \theta)|_{\theta=\theta^\star}$ as the augmented score function evaluated at the true parameter value $\theta^\star$. This is in contrast to the ordinary Fisher score function, as the model now includes the normalizing constant $c$ as one of the parameters. In condition 3, $J_g^\infty(\theta)$ denotes the objective $J_g(\theta)$ from (7) for $N_d, N_n \to \infty$, and we require an uniform convergence in $\theta$ of the sample version $J_g(\theta)$ towards it.

Theorem 2 establishes that the parameterized sample version of our estimator has the same desirable properties as in the non-parametric case in theorem 1. The proof follows closely the corresponding proof of consistency of the Maximum Likelihood estimator. Conditions *1* and *2* are required to make our estimator well defined and are easy to satisfy with proper selection of the auxiliary distribution and the nonlinearities. Condition *3* has its counterpart in Maximum Likelihood estimation where we need the sample version of Kullback-Leibler divergence to converge to the true Kullback-Leibler divergence uniformly over $\theta$ (Wasserman, 2004). Similarly, the full-rank requirement of matrix $\mathcal{I}$ in condition *4* corresponds to the requirement of model identifiability in Maximum Likelihood estimation. Lastly, we need to impose the integrability condition for $\mathcal{I}$, the second part of condition *4*. This comes from the interplay of choices of the auxiliary distribution $p_n$ and the nonlinearities $g_1(\cdot)$ and $g_2(\cdot)$.

Having established the consistency of our estimator we will now go on to show that it is asymptotically normally distributed, and give an expression for the asymptotic covariance matrix of the estimator. This is of interest since the covariance depends on the choice of the nonlinearities $g_1(\cdot)$ and $g_2(\cdot)$, and the auxiliary distribution $p_n$. The following result can thus guide us on the choice of these design parameters.

**Theorem 3.** (Asymptotic normality) *Given that the conditions from Theorem 2 for $g_1(\cdot)$, $g_2(\cdot)$ and $p_n$ hold, then $\sqrt{N_d}(\hat{\theta}_g - \theta^\star)$ is asymptotically normal with mean zero and covariance matrix*

$$\Sigma_g = \mathcal{I}^{-1} \left[ \int p_d \left(\frac{\gamma p_d + p_n}{p_n}\right) g_2'\left(\frac{p_d}{p_n}\right)^2 \psi \psi^T - (1+\gamma) \left(\int p_d\, g_2'\left(\frac{p_d}{p_n}\right) \psi\right) \left(\int p_d\, g_2'\left(\frac{p_d}{p_n}\right) \psi\right)^T \right] \mathcal{I}^{-1} \quad (8)$$

*where $\mathcal{I}$ was defined in theorem 2, and $\gamma = N_d/N_n$ denotes the ratio of data and auxiliary noise sample sizes.*

We immediately notice that the asymptotic covariance matrix can be divided into two parts, one depending linearly on $\gamma$ and another completely independent of it. This property is exploited later in section 3.3, where we consider how many data and noise samples one should optimally use. Furthermore, we have the following results for interesting special cases.

**Corollary 1.** *If the auxiliary distribution $p_n$ equals the data distribution $p_d$, the asymptotic covariance of $\hat{\theta}_g$ does not depend on the choice of nonlinearities $g_1(\cdot)$ and $g_2(\cdot)$.*

If we assume in addition that the normalizing constant $c$ is not part of the parameter vector $\theta$, we can see an illuminating connection to ordinary Maximum Likelihood estimation. In this case the score function $\psi$ becomes the Fisher score function. Correspondingly, $\mathcal{I}$ becomes proportional to the Fisher information matrix $\mathcal{I}_F$,

$$\mathcal{I}_F = \int p_d \psi \psi^T. \quad (9)$$

Now as the expectation of the Fisher score function is zero at the true parameter value, the right hand side of the term inside the square brackets in (8) vanishes, and we are left with

$$\Sigma_g = (1+\gamma)\mathcal{I}_F^{-1}. \quad (10)$$

From this we obtain the following result

**Corollary 2.** *If the auxiliary distribution $p_n$ equals the data distribution $p_d$, and the normalizing constant $c$ is* not *included in the parameters, then the asymptotic covariance of $\hat{\theta}_g$ equals $(1+\gamma)\mathcal{I}_F^{-1}$, which is $(1+\gamma)$ times the Cramér-Rao lower bound for consistent estimators.*

This result is intuitively appealing. In ordinary MLE estimation, the normalizing constant is assumed to be known exactly, and the random error arises from the fact that we only have a finite sample from $p_d$. In contrast, here we need another sample from the auxiliary density to approximate the integral, which also contributes to the error. In the case of $p_n = p_d$ and equal sample size $N_d = N_n$, we achieve two times the Cramér-Rao bound, as both samples equally contribute to the error. Letting the relative amount of noise grow without bounds, $\gamma$ goes to zero and we retain the inverse of the Fisher information matrix as the covariance matrix $\Sigma_g$. This corresponds to the situation where the infinite amount of noise samples allows us to compute the partition function integral to arbitrary accuracy. It is to be noted that the same phenomenon happens even when the auxiliary density does not necessarily equal the data density, but only with one particular choice of $g_1(\cdot)$, namely $g_1(q) = \log q$. In this case we have essentially reduced our estimator to the ordinary Maximum Likelihood method.

In the following, instead of the full covariance matrix $\Sigma_g$, we will use the mean squared error (MSE) of the estimator to compare the performance of different instances from the family. The MSE is defined as the trace of the asymptotic covariance matrix

$$E_d \|\hat{\theta}_g - \theta^\star\|^2 = \operatorname{tr}(\Sigma_g)/N_d + O(N_d^{-2}). \quad (11)$$

Asymptotically the MSE thus behaves like $\operatorname{tr}(\Sigma_g)/N_d$.

## 3 DESIGN PARAMETERS OF THE ESTIMATOR FAMILY

Our estimator family has essentially three design parameters - the nonlinearities $g(\cdot)$, the auxiliary distribution $p_n$ and the amount of data and noise samples used. In this section we will consider each of these in turn.

### 3.1 CHOICE OF NONLINEARITIES

In the following we will denote the ratio $p_m(\theta)/p_n$ by $q$. The objective functions and their gradients for all choices of $g_1(\cdot)$ and $g_2(\cdot)$ introduced here can be found in Table 1.

#### 3.1.1 Importance Sampling (IS)

If we set

$$g_1(q) = \log q \quad \text{and} \quad g_2(q) = q, \quad (12)$$

we recover the parametric version of the objective function in (2). The resulting estimator is closely related to the Maximum Likelihood Monte Carlo method of Geyer and Thompson (1992), which uses Importance Sampling to handle the partition function. Our objective function is slightly different from theirs due to the fact that the normalizing constant $c$ is estimated as a model parameter.

The gradient of this objective, $J_{IS}(\theta)$ (Table 1, row 1), depends on the ratio $q = p_m(\theta)/p_n$, which can make the method very unstable if $p_n$ is not well matched to the true data density. This is a well known shortcoming of the Importance Sampling method.

#### 3.1.2 Inverse Importance Sampling (InvIS)

An interesting choice is given by setting

$$g_1(q) = -\frac{1}{q} \quad \text{and} \quad g_2(q) = \log(q), \quad (13)$$

which can be considered a reversed version of the Importance Sampling type of estimator above. Here we have moved the logarithm to the second term, while the first term becomes linear in $1/q = p_n/p_m(\theta)$. This inverse ratio can get large if the auxiliary density $p_n$ has a lot of mass at the regions where $p_m(\theta)$ is small. However, this rarely happens as soon as we have a reasonable estimate of $\theta$, since the ratio is evaluated at the points sampled from $p_d$, which are likely to be in the regions where the values of the model $p_m(\theta)$ are not extremely small. Thus the gradient is considerably more stable than in case of $J_{IS}$, especially if the auxiliary density has thinner tails than the data. Furthermore, the form of the second term in the gradient might enable an exact computation of the integral instead of sampling from $p_n$ with some models, such as fully visible Boltzmann Machines (Ackley et al., 1985), which is closely related to Mean Field approximation with certain choice of $p_n$.

#### 3.1.3 Noise Contrastive Estimation (NC)

We get a particularly interesting instance of the family by setting

$$g_1(q) = \log\left(\frac{q}{1+q}\right) \quad \text{and} \quad g_2(q) = \log(1+q). \quad (14)$$

By rearranging we obtain the objective function

$$J_{NC}(\theta) = \int p_d \log \frac{1}{1 + \exp\left(-\log \frac{p_n}{p_m(\theta)}\right)}$$
$$+ \int p_n \log \frac{1}{1 + \exp\left(-\log \frac{p_m(\theta)}{p_n}\right)}, \quad (15)$$

which was proposed in (Gutmann and Hyvärinen, 2010) to estimate unnormalized models. The authors call the resulting estimation procedure Noise Contrastive Estimation. They related this objective function to the log-likelihood in a nonlinear logistic regression model which discriminates the observed sample of $p_d$ from the noise sample of the auxiliary density $p_n$. A connection between density estimation and classification has been made earlier by Hastie et al. (2009).

| Name | $g_1(q)$ | $g_2(q)$ | Objective $J_g(\theta)$ | $\nabla_\theta J_g(\theta)$ |
|------|----------|----------|-------------------------|------------------------------|
| IS | $\log q$ | $q$ | $\mathrm{E}_d \log p_m - \mathrm{E}_n \frac{p_m}{p_n}$ | $\mathrm{E}_d \psi - \mathrm{E}_n \frac{p_m}{p_n}\psi$ |
| PO | $q$ | $\frac{1}{2}q^2$ | $\mathrm{E}_d \frac{p_m}{p_n} - \mathrm{E}_n \frac{1}{2}\left(\frac{p_m}{p_n}\right)^2$ | $\mathrm{E}_d \frac{p_m}{p_n}\psi - \mathrm{E}_n \left(\frac{p_m}{p_n}\right)^2 \psi$ |
| NC | $\log(\frac{q}{1+q})$ | $\log(1+q)$ | $\mathrm{E}_d \log(\frac{p_m}{p_m+p_n}) + \mathrm{E}_n \log(\frac{p_n}{p_m+p_n})$ | $\mathrm{E}_d(\frac{p_n}{p_m+p_n})\psi - \mathrm{E}_n(\frac{p_m}{p_m+p_n})\psi$ |
| InvPO | $-\frac{1}{2q^2}$ | $-\frac{1}{q}$ | $-\mathrm{E}_d \frac{1}{2}\left(\frac{p_n}{p_m}\right)^2 + \mathrm{E}_n \left(\frac{p_n}{p_m}\right)$ | $\mathrm{E}_d \left(\frac{p_n}{p_m}\right)^2 \psi - \mathrm{E}_n \frac{p_n}{p_m}\psi$ |
| InvIS | $-\frac{1}{q}$ | $\log q$ | $-\mathrm{E}_d \frac{p_n}{p_m} - \mathrm{E}_n \log p_m$ | $\mathrm{E}_d \frac{p_n}{p_m}\psi - \mathrm{E}_n \psi$ |

Table 1: Objective functions and their gradients for the different choices of nonlinearities $g_1(\cdot)$ and $g_2(\cdot)$

Here we show that Noise Contrastive Estimation can be seen as a special case of the larger family of density estimation methods.

Table 1 shows that in the gradient of the Noise Contrastive estimator, the score function $\psi$ is multiplied by a ratio that is always smaller than one. This indicates that the gradient is very stable.

### 3.1.4 Polynomial (PO) and Inverse Polynomial (InvPO)

More examples of nonlinearities are given by polynomials and rational functions. We consider here one with a second degree polynomial in the numerator, and one with a second degree polynomial in the denominator. These are given by

$$g_1(q) = q \quad \text{and} \quad g_2(q) = \frac{1}{2}q^2 \qquad (16)$$

and

$$g_1(q) = -\frac{1}{2q^2} \quad \text{and} \quad g_2(q) = \frac{1}{q}, \qquad (17)$$

respectively.

The proposed nonlinearities are recapitulated in Table 1. Simulations in section 4 will investigate which nonlinearities perform well in different estimation tasks.

### 3.2 CHOICE OF AUXILIARY DISTRIBUTION

In selecting the auxiliary distribution $p_n$, we would like it to fulfill at least the following conditions: *1)* It should be easy to sample from, *2)* we should be able to evaluate the expression of $p_n$ easily for the computation of the gradients and *3)* it should lead to small MSE of the estimator.

Finding the auxiliary distribution which minimizes the MSE is rather difficult. However, it is instructive to look at the Importance Sampling estimator in the first row of table 1. In this case we can obtain a formula for the optimal $p_n$ in closed form.

**Theorem 4.** (Optimal $p_n$) Let $g_1(q) = \log q$ and $g_2(q) = q$ so that the objective function becomes $J_{IS}(\theta)$. Then the density of the auxiliary distribution $p_n(\mathbf{u})$ which minimizes the MSE of the estimator is given by

$$p_n(\mathbf{u}) \propto ||\mathcal{I}^{-1}\psi(\mathbf{u})||\, p_d(\mathbf{u}) \qquad (18)$$

where $\mathcal{I}$ was defined in theorem (2).

This tells us that the auxiliary density $p_n$ should be the true data density $p_d$, scaled by a norm of something akin to the natural gradient (Amari, 1998) of the log-model $\mathcal{I}^{-1}\psi$ evaluated at the true parameter values. Note that this is different from the optimal sampling density for traditional Importance Sampling (see e.g. Wasserman, 2004), which usually tries to minimize the variance in the estimate of the partition function integral alone. Here in contrast, we aim to minimize the MSE of all the model parameters at the same time.

Choosing an MSE minimizing auxiliary density $p_n$ is usually not attractive in practice, as it might not be easy to sample from especially with high dimensional data. Furthermore, the theorem above showed that we need to know the true data density $p_d$, which we are trying to estimate in the first place. Hence we think it is more convenient to use simple auxiliary distributions, such as Gaussians, and control the performance of the estimator by appropriate choices of the nonlinearities.

### 3.3 CHOICE OF THE AMOUNT OF NOISE

Recall that we used $\gamma$ in theorem 3 to denote the ratio $N_d/N_n$. Also note that $\gamma$ goes to zero as the amount of noise samples $N_n$ grows to infinity. Let $N_{tot} = N_d + N_n$ denote the total amount of samples. In our simulations, the computation time increases approximately linearly with $N_{tot}$. Thus we can use $N_{tot}$ as a proxy for the computational demands.

Given a fixed number of samples $N_d$ from $p_d$, the form of the covariance matrix $\Sigma_g$ tells us that increasing the amount of noise always decreases the asymptotic variance, and hence the MSE. This suggests that we should use the maximum amount of noise given the computational resources available. However, assuming that we can freely choose how much data and noise we use, it becomes compelling to ask what the optimal ratio $\hat{\gamma}$ is, given the available computational resources $N_{tot}$.

Asymptotically we can write the MSE of the estimator as
$$\mathrm{E}_d\,||\,\hat{\theta}_g - \theta^\star||^2 = \frac{1+\gamma^{-1}}{N_{tot}}\mathrm{tr}(\Sigma_g) \qquad (19)$$

which enables us to find the ratio $\hat{\gamma}$ that minimizes the error given $N_{tot}$. This is easy, as the expression for $\Sigma_g$ breaks down to two parts, one linear, and the other not depending on $\gamma$. Minimization gives us

$$\hat{\gamma} = \arg\min_\gamma \frac{1+\gamma^{-1}}{N_{tot}}\mathrm{tr}(\Sigma_g)$$
$$= \left(\frac{\mathrm{tr}\left[\mathcal{I}^{-1}\left[\mathcal{A}-\mathcal{B}\right]\mathcal{I}^{-1}\right]}{\mathrm{tr}\left[\mathcal{I}^{-1}\left[\mathcal{A}_\gamma-\mathcal{B}\right]\mathcal{I}^{-1}\right]}\right)^{\frac{1}{2}} \qquad (20)$$

where
$$\mathcal{A}_\gamma = \int p_d \frac{p_d}{p_n} g'_2\left(\frac{p_d}{p_n}\right)^2 \psi\psi^T \qquad (21)$$
$$\mathcal{A} = \int p_d\, g'_2\left(\frac{p_d}{p_n}\right)^2 \psi\psi^T \qquad (22)$$
$$\mathcal{B} = \left(\int p_d\, g'_2\left(\frac{p_d}{p_n}\right)\psi\right)\left(\int p_d\, g'_2\left(\frac{p_d}{p_n}\right)\psi\right)^T \qquad (23)$$

and $\mathcal{I}$ is as in theorem (2). Here the $\mathcal{A}_\gamma$ is the part of the covariance matrix $\Sigma_g$ which depends linearly on $\gamma$.

## 4 SIMULATIONS

We will now illustrate the theoretical properties of the estimator family derived above. We use it to estimate the mixing matrix and normalizing constant of a 4 dimensional ICA model (Hyvärinen et al., 2001). The data $\mathbf{x} \in \mathbb{R}^4$ is generated by

$$\mathbf{x} = \mathbf{A}\mathbf{s}, \qquad (24)$$

where $\mathbf{A}$ is a four-by-four mixing matrix chosen at random. All four sources $\mathbf{s}$ are independent and follow a generalized Gaussian distribution of unit variance and zero mean. The data and model density are given by

$$p_d(\mathbf{x}) = \frac{\det |\mathbf{B}^\star|}{\left(\kappa(\alpha)\nu(\alpha)\right)^d}\exp\left(-\left|\left|\frac{\mathbf{B}^\star\mathbf{x}}{\nu(\alpha)}\right|\right|_\alpha^\alpha\right) \qquad (25)$$

$$p_m(\mathbf{x};\theta) = \exp\left(-\left|\left|\frac{\mathbf{B}\mathbf{x}}{\nu(\alpha)}\right|\right|_\alpha^\alpha + c\right) \qquad (26)$$

where $\alpha > 0$ is a shape parameter, $d$ is the dimension of the model and

$$\kappa(\alpha) = \frac{2}{\alpha}\Gamma\left(\frac{1}{\alpha}\right) \quad \text{and} \quad \nu(\alpha) = \sqrt{\frac{\Gamma\left(\frac{1}{\alpha}\right)}{\Gamma\left(\frac{3}{\alpha}\right)}}. \qquad (27)$$

The Gaussian distribution is recovered by setting $\alpha = 2$. We used $\alpha = 1$ (Laplacian-distribution) and $\alpha = 3$ for simulations with super- and sub-Gaussian sources, respectively. The parameters $\theta \in \mathbb{R}^{17}$ consists of the mixing matrix $\mathbf{B}$ and an estimate of the negative log-normalizing constant $c$. True parameter values are denoted as $\mathbf{B}^\star = \mathbf{A}^{-1}$. The auxiliary distribution $p_n$ is a multivariate Gaussian with the same covariance structure as the data, and the model is learned by gradient ascent on the objective function $J_g(\theta)$ defined in (7). For the optimization, we used a standard conjugate gradient algorithm (Rasmussen, 2006).

Figure 1 shows the estimation results for the super-Gaussian model for the nonlinearities in table 1. The figure shows that all estimators for which the integrability condition in theorem 2 hold, namely NCE, InvIS, InvPO, have a MSE that decreases linearly in $N_d$. This confirms the consistency result from theorem 2. Both NC and InvIS perform better than InvPO, and of these two NC is slightly better. The estimator IS is not consistent, but still gives reasonable results for a finite sample size. For the sub-Gaussian case, the best performing nonlinearity was NC, while IS performed this time better than InvIS (results not shown).

The practical utility of the Noise Contrastive objective function in estimation of multi-layer and Markov Random Field-models has been previously demonstrated by Gutmann and Hyvärinen (2010) who also showed that it is more efficient than traditional Importance Sampling, Contrastive Divergence (Hinton, 2002) and Score Matching (Hyvärinen, 2005).

We also numerically fit the optimal nonlinearity for the ICA model using an orthogonal polynomial basis to construct $g'_2(\cdot)$ and minimized the asymptotic MSE with respect to the coefficients of the basis functions. The derivative of the optimal $g_2(\cdot)$ is plotted in Figure 2 both in sub- and super-Gaussian case, along with the corresponding derivatives of the nonlinearities from table 1. Interestingly, for the super-Gaussian model, Noise Contrastive nonlinearity seems to be particularly close to the optimal one, whereas in the sub-Gaussian case, the optimum is somewhere between Importance Sampling and Noise Contrastive nonlinearities.

Figure 2 also nicely illustrates the trade-off between stability and efficiency. If $p_n$ is not well matched to the data density $p_d$, the ratio $q = p_m(\theta)/p_n$ can get extremely large or extremely small. The first case is especially problematic for Importance Sampling (IS). To remedy this, we need $g_2'(\cdot)$ to decay fast enough, else the estimation becomes unstable. However, decaying too fast means that we do not use the available information in the samples efficiently. In the second case, estimators like Inverse Importance Sampling (InvIS) have problems since the $g_2'(\cdot)$ grows without bounds at zero. Thus $g_2'$ should be bounded at zero. The Noise Contrastive nonlinearity seems to strike a good balance between all these requirements.

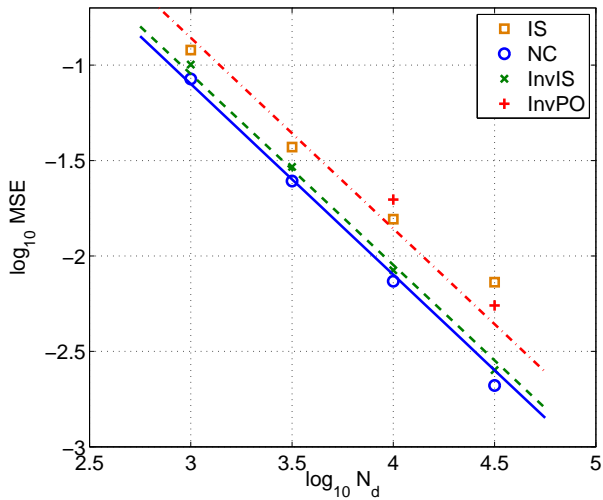
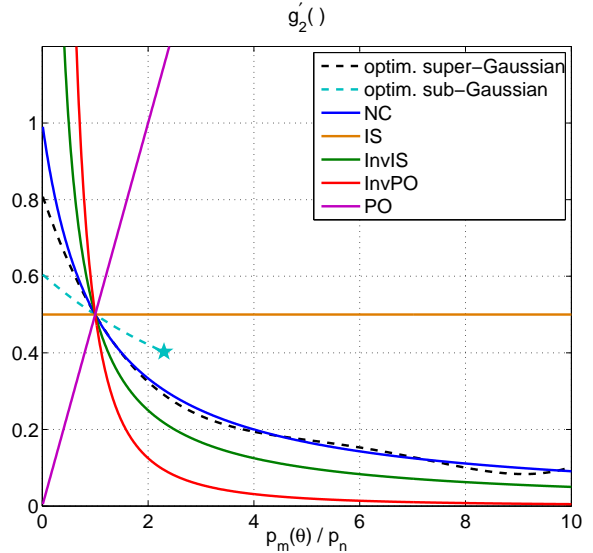

Figure 1: MSE from the simulations of the super-Gaussian ICA model specified in section 4, with equal amounts of data and noise samples and a Gaussian density with the covariance structure of the data as $p_n$. MSE was computed together for parameters $\varphi$ and normalizing constant $c$. Different colors denote different choices of the nonlinearities. Lines show the theoretical predictions for asymptotic MSE, based on theorem 3. Note that even though the asymptotic MSE for the IS nonlinearity is infinite for this model, it seems to perform acceptably with finite sample sizes. The optimization with the PO nonlinearity did not converge. This can be understood from the gradient given in table 1. The term $(p_m/p_n)^2$ gets extremely large when the model density has fatter tails than the auxiliary density $p_n$. For the NC, InvIS and InvPO nonlinearities the MSE goes down linearly in $N_d$, which validates the consistency result from theorem 2.

Figure 2: Numerically fitted optimal nonlinearities for the sub- and super-Gaussian ICA model with Gaussian $p_n$ are plotted with dashed lines. For the super-Gaussian case we cut the ratio $p_m(\theta)/p_n$ at 10, so that only few samples were rejected. For the sub-Gaussian case the ratio is bounded, and the optimal $g_2'(\cdot)$ was fitted only up to the maximum value around 2.2 (marked with $\star$). The solid lines correspond to the different nonlinearities from Table 1. The x-axis is the argument of the nonlinearity, i.e. the ratio $q = p_m(\theta)/p_n$. As the objective functions can be multiplied by a non-zero constant, also MSE is invariant under the multiplication of $g_2'(\cdot)$. Thus all the nonlinearities were scaled to match at $q = 1$.

Finally, we computed the MSE for the different nonlinearities with different choices of the ratio $\gamma = N_d/N_n$, assuming that the computational resources $N_{tot}$ are kept fixed. The results for the super-Gaussian case are shown in Figure 3. The optimal ratio $\hat{\gamma}$ computed in section 3.3 is also shown. It varies between the nonlinearities, but is always relatively close to one.

## 5 CONCLUSIONS

We introduced a family of consistent estimators for unnormalized statistical models. Our formulation of the objective function allows us to estimate the normalizing constant just like any other model parameter. Because of consistency, we can asymptotically recover the true value of the partition function. The explicit estimate of the normalizing constant $c$ could thus be used in model comparison.

Our family includes Importance Sampling as a special case, but depending on the model, many instances perform superior to it. More importantly, the performance of certain nonlinearities, such as the ones in Noise Contrastive Estimation, is robust with respect to the choice of the auxiliary distribution $p_n$, since the both parts of the gradient remain bounded (see Table 1). This holds independent from the characteristics of the data, which makes this method applicable in a wide variety of estimation tasks.

Many current methods rely on MCMC-sampling for approximating the partition function. We emphasize that our method uses a single sample from a density, which we can choose to be something convenient such as a multivariate Gaussian. The objective functions do not have integrals or other difficult expressions that might be costly to evaluate. Furthermore, the form of the objectives allows us to use back-propagation to efficiently compute gradients in multi-layer networks.

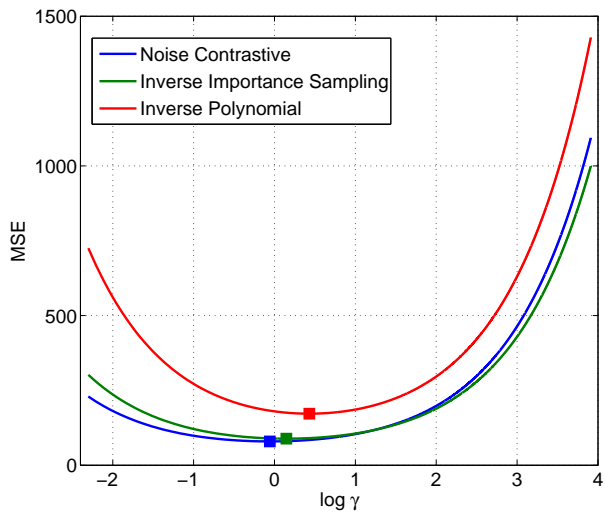

Figure 3: MSE of the estimator as a function of the ratio $\gamma = N_d/N_n$ plotted on a natural logarithmic scale. Here we assume that $N_{tot} = N_d + N_n$ is kept fixed. The different curves are for the different nonlinearities $g(\cdot)$ (see Table 1). The squares mark the optimal ratio $\hat{\gamma}$ for a given estimator. These results are computed for the super-Gaussian ICA model given in section 4 using Gaussian noise with the covariance structure of data as $p_n$. Note that the asymptotic MSE is not finite for IS and PO nonlinearities for any $\gamma$ under this model, as they violate the integrability condition 4. in theorem 2.

The objective function is typically smooth so that we can use any out-of-the-shelf gradient algorithm, such as conjugate gradient, or some flavor of quasi-Newton for optimization. It is also a question of interest if some kind of efficient approximation to the natural gradient algorithm could be implemented using the metric defined by $\mathcal{I}$. Furthermore, having a well defined objective function permits a convenient analysis of convergence as opposed to the Markov-Chain based methods such as Contrastive Divergence (Hinton, 2002).

Finally, our method has many similarities to robust statistics (Huber, 1981), especially M-estimators, which are used in the estimation of location and scale parameters from data with gross outliers. Our approach differs from this, however, since here the need for robustness arises not so much from the characteristics of the data, but from the auxiliary density that we needed to introduce in order to make the estimation of unnormalized models possible. Our future work is aimed to elucidate this connection further.